\documentclass{article}
\usepackage{amsmath}
\usepackage[draft]{hyperref}
\usepackage{amsfonts}
\usepackage{amssymb}
\usepackage[english]{babel}
\usepackage{graphicx}
\usepackage{natbib}
\usepackage{wrapfig}
\usepackage{float}

     \usepackage[preprint]{template}

\usepackage{listings}
\usepackage{comment}
\usepackage{xcolor}
\usepackage{tcolorbox}

\usepackage[utf8]{inputenc} 
\usepackage[T1]{fontenc}    
\usepackage{url}            
\usepackage{booktabs}       
\usepackage{amsfonts}       
\usepackage{nicefrac}       
\usepackage{microtype}      
\usepackage{xcolor}         
\usepackage{subcaption}

\title{Highly Parallelized Reinforcement Learning Training\\
	with Relaxed Assignment Dependencies}

\usepackage{authblk}

\author{
	Zhouyu He\textsuperscript{\rm 1,\rm 2, \rm $\star$}, Peng Qiao\textsuperscript{\rm 1, \rm 2,\rm $\star$}, Rongchun Li\textsuperscript{\rm 1,\rm 2, \rm $\dagger$}, Yong Dou\textsuperscript{\rm 1,\rm 2},  Yusong Tan\textsuperscript{\rm 1,\rm 2,\rm $\dagger$} \\
  \textsuperscript{\rm 1} College of Computer Science and Technology, National University of Defense Technology \\
  \textsuperscript{\rm 2} National Key Laboratory of Parallel and Distributed Computing, National University of Defense Technology \\
  \textsuperscript{\rm $\star$} These authors contributed equally to this work.\\
  \textsuperscript{\rm $\dagger$} Corresponding authors: \{rongchunli, ystan\}@nudt.edu.cn\\
}

\usepackage{bibentry}
\usepackage{pifont}
\usepackage{amsmath}
\usepackage{graphicx}
\usepackage{algorithm}
\usepackage{algorithmic}
\usepackage{color}
\usepackage{caption}
\usepackage{appendix}
\usepackage{multirow}

\begin{document}

\maketitle

\begin{abstract}
As the demands for superior agents grow, the training complexity of Deep Reinforcement Learning (DRL) becomes higher. Thus, accelerating training of DRL has become a major research focus. Dividing the DRL training process into subtasks and using parallel computation can effectively reduce training costs. However, current DRL training systems lack sufficient parallelization due to data assignment between subtask components. This assignment issue has been ignored, but addressing it can further boost training efficiency. Therefore, we propose a high-throughput distributed RL training system called TianJi. It relaxes assignment dependencies between subtask components and enables event-driven asynchronous communication. Meanwhile, TianJi maintains clear boundaries between subtask components. To address convergence uncertainty from relaxed assignment dependencies, TianJi proposes a distributed strategy based on the balance of sample production and consumption. The strategy controls the staleness of samples to correct their quality, ensuring convergence. We conducted extensive experiments. TianJi achieves a convergence time acceleration ratio of up to 4.37 compared to related comparison systems. When scaled to eight computational nodes, TianJi shows a convergence time speedup of 1.6 and a throughput speedup of 7.13 relative to XingTian, demonstrating its capability to accelerate training and scalability. In data transmission efficiency experiments, TianJi significantly outperforms other systems, approaching hardware limits. TianJi also shows effectiveness in on-policy algorithms, achieving convergence time acceleration ratios of 4.36 and 2.95 compared to RLlib and XingTian. TianJi is accessible at
\href{https://github.com/HiPRL/TianJi.git}{https://github.com/HiPRL/TianJi.git}.

\end{abstract}

\section{Introduction}
Reinforcement learning(RL) is a powerful approach for addressing decision-making problems. An agent is trained to
interact with the environment based on its policy to maximize long-term rewards \cite{li2017deep,chadi2023understanding}. To address complex spatial problems, value function approximation methods using neural networks are commonly employed \cite{sutton1999policy,grondman2012survey,schulman2015trust,wilcox2022monte}. Deep Reinforcement Learning (DRL) combines the strong expressive
power of deep learning with the decision-making capabilities of RL, offering a versatile system for flexible decision-making and control\cite{dqn,ddpg,ppo}. Research and applications of DRL
are growing more complex\cite{brittain2024improving,wang2023dm2,weihs2020allenact,szot2023large,chen2023end,jiang2024rocket}. In complex scenarios,
the scale of the state and action spaces is vast, and computational complexity grows exponentially\cite{kakade2003sample,neumann2022scaling,li2023deep,Liu2022HeterogeneousSL,RiskQ}. Modern RL methods require substantial computational resources\cite{alphago,alphastar,pala,asurvey}. For example, DQN\cite{dqn} needed continuous training for 12 to 14 days on a single game, involving millions of interactions. OpenAI Five\cite{openaifive} utilized 256 P100 GPUs and trained for 10 months. Due to these immense computational
demands, RL research has increasingly focused on accelerating training.

Parallelization accelerates training by decomposing computational tasks and executing them simultaneously across multiple computing devices or even nodes\cite{isard2007dryad,dean2012large,9428188,10077903}. The typical RL training process can be grouped into three subtasks, as \ding{192}\ding{193}\ding{194} shown in Algorithm
1. These subtasks can be parallelized either individually or
in combination. However, these parallel subtasks must be
executed in a sequential order considering the assignment
dependencies between them, as I-III shown in Algorithm 1.
When these subtasks are assigned into different abstraction
which we will discuss in the following, the assignment dependencies impair the parallel performance. Reasonably relaxing the limitations caused by the above assignment dependencies is the key to accelerating training.

Therefore, the previous works of accelerating the DRL
training can be categorized into four groups. In Figure 1,
we show the abstraction, data dependencies, communication, execution order, and resource utilization of various RL training systems. The Gorila-style architecture\cite{nair2015massively,liang2018rllib,pan2022optimizing,zhu2023msrl,mei2023srl} abstracts sampling-related tasks \ding{192} and \ding{193} as actors and task \ding{194} as a learner, enabling parallelization by component. Gorila-style achieves parallelization of equivalent serial execution logic while preserving data dependencies (I, II, III). These dependencies result in process waiting, as evidenced by the numerous idle blocks in Figure 1(b). A detailed analysis of other architectures is
provided in Section 2.2. Existing systems exhibit significant inter-component dependencies, which severely hinder performance.

\begin{algorithm}[t]
	\caption{Pseudo-code for Function Approximation-based Temporal Difference(0)}
	\begin{algorithmic}[1]
		\small
		\renewcommand{\algorithmicrequire}{\textbf{Input:}}
		\REQUIRE exploration rate $\epsilon$, discount factor $\gamma$, network update rate $\alpha$, value function $V$
		\renewcommand{\algorithmicensure}{\textbf{Output:}}
		\ENSURE network parameters $\theta$
		\STATE Initialize environment, initialize state $s$.
		\FOR{each episode}
		\WHILE{$s$ is not terminal}
		\STATE Get state $s$ and weights $\theta$. \\
		\textcolor{red}{\COMMENT{I: Assignment dependence from \ding{194} to \ding{192}.}}
		\STATE Choose action $a$ with $\epsilon$-greedy policy based on $V(s, \theta)$. 
		\textcolor{blue}{\COMMENT{\ding{192}: Policy inference.}}
		\STATE Get action $a$. \\
		\textcolor{red}{\COMMENT{II: Assignment dependence from \ding{192} to \ding{193}.}}
		\STATE Take action $a$, get reward $r$ and next state $s'$, $s = s'$. \\
		\textcolor{blue}{\COMMENT{\ding{193}: Simulation execution.}}
		\STATE Get trajectory $(s, a, r, s')$. \\
		\textcolor{red}{\COMMENT{III: Assignment dependence from \ding{193} to \ding{194}.}}
		\STATE Compute the TD error:\\
		$\delta = r + \gamma \times V(s'; \theta) - V(s; \theta)$;\\
		Update $\theta$ using SGD: \\
		$\theta = \theta + \alpha \times \delta \times \nabla V(s; \theta)$. \\
		\textcolor{blue}{\COMMENT{\ding{194}: Policy update.}}
		\ENDWHILE
		\ENDFOR
		\RETURN $\theta$
	\end{algorithmic}
\end{algorithm}

Accelerating RL training is a rapidly evolving field. However, current systems often struggle with limited parallelization. They typically decompose large computational tasks
into smaller subtasks for parallel processing, which leads
to various inter-component assignment dependencies.Our
analysis of parallelizability in RL shows that relaxing these
dependencies is crucial for speeding up training. Existing
work either overlooks or fails to achieve these relaxations(I,
II, III). Therefore, we propose TianJi, a distributed RL training system designed to relax assignment dependencies. Our
main contributions are as follows:

\begin{itemize}
	\item We identified the relaxing of assignment dependencies I-III is crucial for accelerating the DRL training. Therefore, we propose TianJi, a distributed reinforcement learning training system that relaxes these dependencies.
	\item We propose a decentralized, data-driven training model
	that transforms inter-component assignment dependencies into asynchronous, loosely-coupled processes. TianJi defines clear component boundaries with internal self-loop computations and event-driven asynchronous communication between components.
	\item We introduce a distributed training strategy that balances
	the training sample production and consumption. Performance analysis manages sample staleness and adjusts sample distribution, ensuring convergence.
	\item We conducted extensive experiments in which TianJi demonstrated up to a 4.37-fold speedup in convergence time compared to related systems. When scaled to eight computing nodes, TianJi achieved a 1.6-fold improvement in convergence time and a 7.13-fold increase in throughput, highlighting its training acceleration capabilities and scalability.
\end{itemize}

\section{Related Work}
\subsection{Parallelization of Reinforcement Learning}
Most RL methods are variants of the Temporal Difference (TD)\cite{sutton1988learning}. TD is a commonly used method in RL that updates the value function based on each step’s trajectory. The TD($\theta$), shown in Algorithm 1, represents the general RL training process. The agent interacts with the environment through multiple discrete time steps t. At each time step t, the agent observes a state s and selects an action $a$ from a set of possible actions according to the policy $\theta$ (denoted as \ding{192}). The agent then interacts with the environment, which advances the simulator to yield the next state $s'$ and a scalar reward $r$ (denoted as \ding{193}). Steps \ding{192} and \ding{193} are repeated until the agent reaches a terminal state or the specified time step, which is simplified to one step in Algorithm 1. During training, the TD error is calculated from the collected trajectories to learn the state value function under the policy from a series of incomplete episodes (denoted as \ding{194}).

Computation task \ding{192}, \ding{193}, and \ding{194} can be parallelized either individually or in combination. Abstracting these tasks as separate or combined components leads to varying levels of data isolation due to different degrees of computational
isolation. The dependencies between these tasks are as follows: \ding{192} requires the state $s$ from \ding{193} and the latest network parameters $\theta$ from \ding{194}(denoted as I); \ding{193} needs the action $a$ from \ding{192}(denoted as II); and  \ding{194} requires the trajectories collected from \ding{193} (denoted as III). When parallelizing, it is crucial to consider the constraints imposed by these dependencies. Therefore, relaxing these dependencies is key to achieving high levels of parallelism.
Shen\cite{10124081} indicated that the parallelism and asynchrony of A3C accelerate convergence in theory. GALA further relaxes the model dependency among workers, resulting in faster convergence than the synchronized A2C. Inspired by previous theoretical and experimental works, we relax assignment dependencies to better utilize computational resources. 

\begin{figure}[!htbp]
	\centering
	\includegraphics[width=0.5\columnwidth]{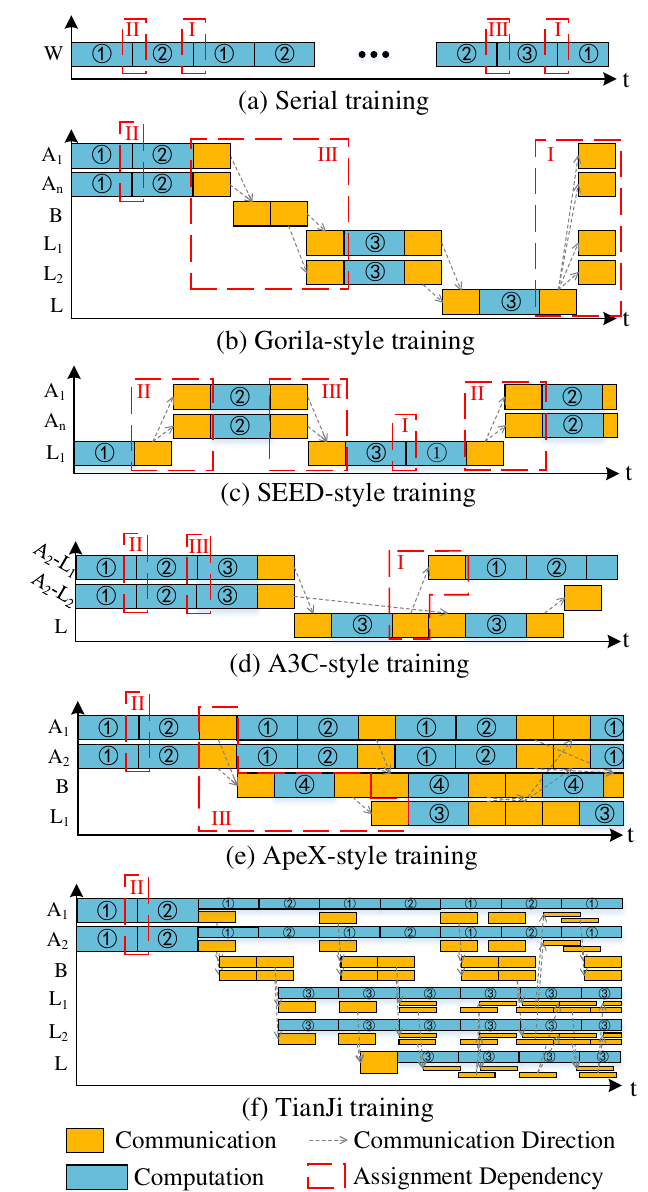} 
	\caption{The spatiotemporal diagram of typical DRL systems during training. The annotations match those in Algorithm 1, with \ding{195} representing the computation of Prioritized Experience Replay (PER). W denotes a worker, A an actor, B a buffer, and L a learner. The A2-L1 represents a component comprising two actors and one learner. This diagram illustrates the abstraction, assignment dependencies, communication patterns, execution sequence, and resource utilization in systems.}
	\vspace*{-15pt}
	\label{fig1}
\end{figure}

\subsection{Parallelization of Existing DRL Training Systems}
The Gorila-style architecture\cite{nair2015massively,liang2018rllib,pan2022optimizing,zhu2023msrl,mei2023srl} abstracts sampling-related \ding{192} and \ding{193} as actors, and \ding{194} as learners, enabling parallel execution. This approach isolates sample production from consumption but requires components to exchange samples and parameters, increasing inter-component communication. Gorila contributed to increasing sampling parallelism by decoupling sampling from model training, allowing each to be parallelized independently. Subsequent developments in Gorila-style architectures include XingTian, which addressed communication bottlenecks with an asynchronous communication channel; SRL, which proposed a data flow-based abstraction to improve resource efficiency. Although these improvements optimize sample quality, communication, and task-resource mapping, they do not contribute to enhancing training parallelism. As shown in Figure \ref{fig1}, Gorila-style architectures are constrained by component assignment dependencies(I, II, III), resulting in significant idle periods.

The SEED-style architecture \cite{espeholt2019seed,petrenko2020sample,zhu2023msrl,mei2023srl} identifies inefficiencies in resource utilization present in the Gorila-style architecture.In Gorila, actors alternate between two dissimilar tasks, \ding{192} and \ding{193}. In SEED-style, model-related tasks \ding{192} and \ding{194} are abstracted into a learner, while actors handle \ding{193}. At each environment step, the state is sent to the learner, which infers the action and then returns it to the actor. This introduces a new issue: latency. To address this issue, SEED implements a high-performance gRPC library, and SampleFactory designs dual-buffer sampling method. Similar to Gorila-style, SEED-style architecture also employs equivalent serial execution logic, constrained by component assignment dependencies(I, II, III).

The A3C-style architecture \cite{mnih2016asynchronous,assran2019gossip} abstracts tasks \ding{192}, \ding{193}, and \ding{194} into a single worker, extending computation across multiple workers. Each worker typically uses a Gorila-like architecture internally, with isolated model parameters between workers. A3C employs an Hogwild!-like\cite{recht2011hogwild} asynchronous update method, while GALA uses a ring-based asynchronous update method. A3C-style reduces parameter synchronization dependencies among workers, so dependency III is only related to the global worker or neighboring workers, significantly increasing training throughput. However, assignment dependencies within each worker(I and II) still exist.

The ApeX-style architecture\cite{horgan2018distributed,espeholt2018impala,assran2019gossip} is conceptually similar to Gorila-style but theoretically relaxes dependencies I and III. However, this relaxation introduces convergence uncertainty. At any given time, the policy parameters $\theta$ among multiple actors are inconsistent, as are the policies of the actors and the learner. To address this, ApeX designs Prioritized Experience Replay(PER)\cite{schaul2015prioritized} to correct staleness. However, PER requires data consistency. The global priority segment tree maintenance and
updates prevent asynchronous data transfer, effectively moving dependency III to the Buffer, which limits the acceleration of training throughput.

\section{This Work}
We found that assignment dependencies significantly restrict parallelization. To address this, we proposed TianJi, a distributed reinforcement learning training system that relaxes these dependencies. TianJi abstracts \ding{192} and \ding{193} into a single actor, thereby mitigating the negative impact of II. TianJi introduces decentralized data-driven training, transforming assignment dependencies I and III between components into asynchronous data exchanges (see Section 3.1). Decentralized computing leads to data isolation, resulting in inter-process data transfers. Efficient data transfer is the base of achieving high-throughput training. TianJi implements asynchronous communication with hidden overheads (see Section 3.2). Although relaxing assignment dependencies can introduce data staleness, TianJi addresses this by employing a distributed strategy based on production-consumption balance, ensuring training convergence through sample distribution adjustment. This strategy also facilitates scaling, enabling the system to overcome performance bottlenecks (see Section 3.3).

\begin{figure}[t]
	\centering
	\includegraphics[width=0.65\columnwidth]{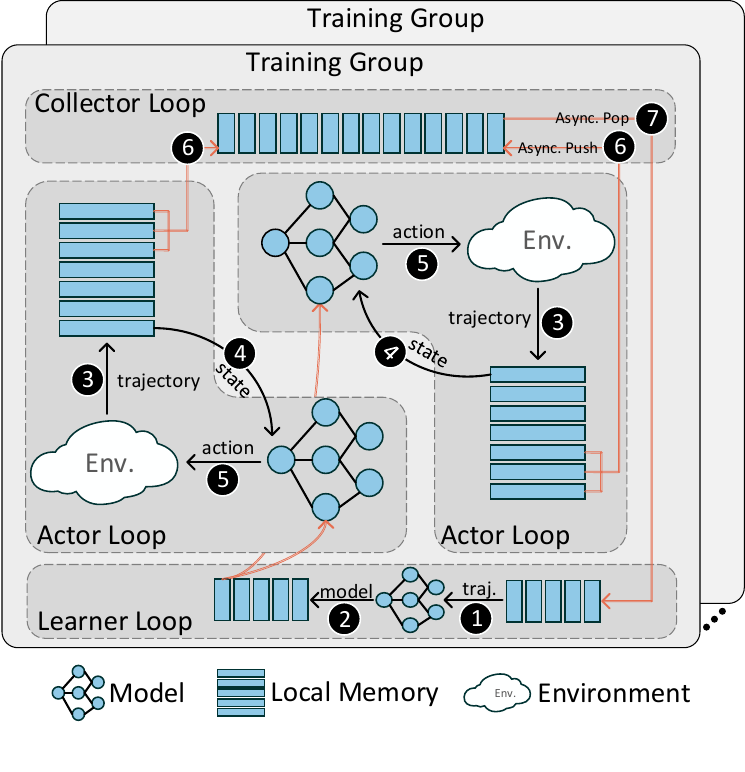} 
	\caption{Data-driven training flowchart with relaxed assignment dependencies.}
	\vspace*{-15pt}
	\label{fig2}
\end{figure}

\subsection{Decentralized Data-driven Training}
Existing systems typically use a global core component to allocate and coordinate multiple instances, ensuring equivalence between the parallel and serial implementations. These implementations maintain assignment dependencies either fully or partially, resulting in significant idle time during training, as illustrated in Figure \ref{fig1}. Therefore, TianJi reorganizes DRL computations to enhance parallelism and introduces a decentralized data-driven training approach. In this approach, assignment dependency II is reduced to intracomponent dependency, while dependencies I and III are converted to asynchronous data exchanges.

TianJi abstracts sampling-related computation tasks (\ding{192}, \ding{193}) as "Actor" and model updates (\ding{194}) as "Learner," parallelizing these at the component level. Figure \ref{fig2} illustrates the training process of TianJi. Once computation begins, each role independently performs looped computations and triggers data exchanges between components at appropriate times. In the Learner Loop, steps (\ding{182}, \ding{183}) are executed repeatedly. Step \ding{182} involves loading trajectories from local storage, using an asynchronous communication trigger to determine if data reception is needed. Step \ding{183} involves learning and updating the model, which is then stored locally, with a trigger to decide whether sending model parameters is necessary. In the Actor Loop, steps (\ding{184}-\ding{186}) are executed in sequence. Step \ding{184} involves the agent interacting with the environment to obtain trajectories or states, which are stored locally, with a trigger to determine if these trajectories should be sent to the buffer. Step \ding{185} involves model inference, with a trigger to decide whether a new model should be received. Step \ding{186} involves generating actions through model inference and sending these actions to the environment. In the Buffer Loop, steps (\ding{187}, \ding{188}) are executed repeatedly to manage asynchronous data reception and sending. TianJi supports scaling the computation across multiple training groups. 

\subsection{Event-driven Asynchronous Communication}
Distributed components create data isolation, necessitating communication between components. Experiments have shown that communication can sometimes take more time than computation\cite{pan2022optimizing,zhao2023high}. We observed that learners do not require samples to be uniformly distributed across actors. Therefore, common practices such as group communication or uniformly requesting data from each actor can be both impractical and costly. TianJi employs an actively pushing asynchronous communication mode, where communication requests are initiated by the sender. Once the data at the sender is ready, it immediately triggers the communication, actively pushing the
data to the intended recipient. The receiver uses a probing mechanism. The sender does not wait for an immediate request and continues its computations. Communication is confirmed during subsequent data transmissions, allowing computation and communication to overlap.

\begin{figure}[t]
	\centering
	\includegraphics[width=0.7\columnwidth]{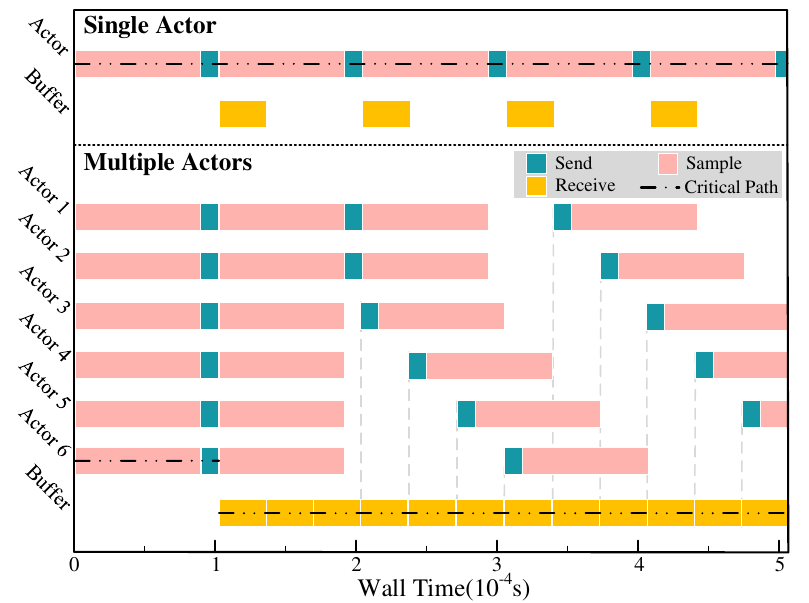} 
	\caption{As the number of actors increases, the critical path
		shifts. Critical path is the sequence of tasks that determines
		the minimum time required to complete the computation.}
	\vspace*{-15pt}
	\label{fig3}
\end{figure}

We will analyze the overlap between communication and computation in TianJi by examining critical path transitions during the sample collection phase. In this phase, the Actor collects data and sends it to the Buffer, where the samples are received. Figure \ref{fig3} illustrates the critical path transitions in both "Single Actor" and "Multiple Actors" scenarios. In the "Single Actor" scenario, the sampling time exceeds the reception time. Causing computation overshadows communication, making the Actor’s sampling the critical determinant of the path. Conversely, in the "Multiple Actors" scenario, increasing the number of actors effectively reduces the sampling time. When the sampling time becomes shorter than the communication time, communication can no longer be
entirely concealed, shifting the key path to the Buffer. Although computation can obscure communication, this concealment has its limitations.

The theoretical collect time $T_c$ can be calculated. When the communication hidden limit is not reached, the communication time can be masked by the computation time, placing the critical path within the "Actor". When the communication hidden limit is reached, the communication time can no longer be masked by the computation time, causing the critical path to shift to the Buffer. From the Buffer’s perspective, increasing the number of actors is equivalent to linearly accelerating the sample collection process. Based on the critical path transition rules, we can derive the following formula:
\begin{equation}
	T_c = 
	\begin{cases} 
		(T_{\text{sp}} + T_{\text{sd}}) \times N_{\text{s}}, & \text{if } \frac{(T_{\text{s}} + T_{\text{sd}})}{N_{\text{A}}} > T_{\text{rv}} \\
		T_{\text{rv}} \times N_{\text{s}}, & \text{others}
	\end{cases}
\end{equation}
where, $N_s$ is the number of collected trajectories, $T_{sp}$ is the
time of a single sampling. $T_{sd}$ and $T_{rv}$ are the time of sending and receiving, respectively. $N_A$ is the number of actors.

Event-driven asynchronous communication eliminates redundant waiting and enables the overlap of computation and communication, thereby facilitating more efficient and rational data exchange between components.

\subsection{Distribution Strategy}
Distributed strategies handle the configuration of components and their allocation to computing resources, impacting the location of performance bottlenecks and resource utilization. When a performance bottleneck arises, efforts outside this bottleneck do not contribute to convergence.
Extensive experiments have demonstrated that performance bottlenecks vary across different algorithms, applications, and hardware. These studies indicate that some algorithms are more prone to bottlenecks in specific areas, resulting in fixed strategies targeting particular performance issues. However, because of the unpredictability of performance bottlenecks, fixed optimization techniques cannot guarantee overall throughput improvement. Reverb\cite{reverb} proposes SPI rate limiting, which controls when items can be inserted into or sampled from a table. However, SPI reflects the ratio over a period of time, varies significantly across different scales, and involves blocking operations. SPI is both inaccurate and inefficient.
Therefore, TianJi proposes a distributed strategy based on performance analysis and production-consumption balance to enable scalable training that addresses performance bottlenecks. This strategy also maintains sample freshness by ensuring consistency between sample distribution and serial execution, which is essential for training convergence.

Given the algorithm, application, and computing resources, the goal of a distributed strategy is to maximize computational throughput and ensure training convergence. TianJi allows the assignment of specific computing resources to each component, which can be easily managed using a configuration file. For example, with a only CPUs setup and a total of M cores, performance analysis tools can measure single-step training to obtain the consumption throughput $TR_L$ and production throughput $TR_A$.

In serial execution, once the sample pool is filled, the sample staleness $P$ stabilizes. The formula is:
\begin{equation}
	P = \frac{B_s}{N_s}
\end{equation}
where, $B_s$ is the size of a batch trajectories, $N_s$ is the size of
buffer.

The optimization objective under the distributed strategy is to maximize training throughput while controlling the sample obsolescence ratio, as follows:
\begin{align}
	\text{Maximize } & \min(TR_L, TR_A) \\
	\text{Subject to: } & \left\{
	\begin{array}{l}
		\frac{N_c}{N_s'} = P \\
		M_L + M_A = M \\
		TR_L \approx TR_A
	\end{array}
	\right.
\end{align}
where, $N_c$ is the number of trajectories consumed per unit time, $M_A$ and $M_L$ are the number of cores obtained by the Learner and the Actor, respectively.

Through performance analysis and objective function optimization, TianJi achieves load balancing and optimal mapping of computational tasks and resources.

\section{Evaluation}
The evaluation answer the following questions: (1) How does TianJi compare to existing systems in terms of overall performance optimization (Section 4.2)? (2) How effective are the key components as demonstrated by the ablation study (Section 4.3)?

\subsection{Setting}
Testbed. We configured two hardware platforms for our experiments. The first platform is a CPU-only Slurm cluster with 8 computing nodes. Each node is equipped with 2 Intel Xeon Gold 6248 processors, providing a total of 40 physical cores per node. Each node has 384GB of memory, and they are interconnected using ConnectX-6 high-speed interconnects. The second platform is a heterogeneous machine, equipped with one A100 GPU, 40 physical cores, and 376GB of memory.

Baseline. The three baselines are: ApeX, a distributed architecture for large-scale deep reinforcement learning that employs prioritized experience replay; RLlib, a widely used DRL training system based on Ray; and XingTian, which leverages communication-computation overlap in DRL to offer a more efficient asynchronous communication channel than other systems.

Algorithms \& Environment. We conducted experiments in Atari and OpenAI Gym. The algorithms evaluated are DQN(off-policy) and PPO(on-policy). We used RLlib’s default network architecture and parameters as benchmarks for both Gym and Atari tasks. All comparisons used identical network architectures and hyperparameters across the same games to ensure fairness.

Metrics. The evaluation focuses on learning throughput and performance. learning throughput: This includes sample throughput, receive throughput, and training throughput, which represent the number of frames processed per second. learning performance: This is assessed using episode return and final time, which indicate the reward value and the time taken to achieve a specified reward, respectively. Reward are
averaged over a sliding window of 100 episodes.

\begin{figure*}[t]
	\centering
	\includegraphics[width=1\columnwidth]{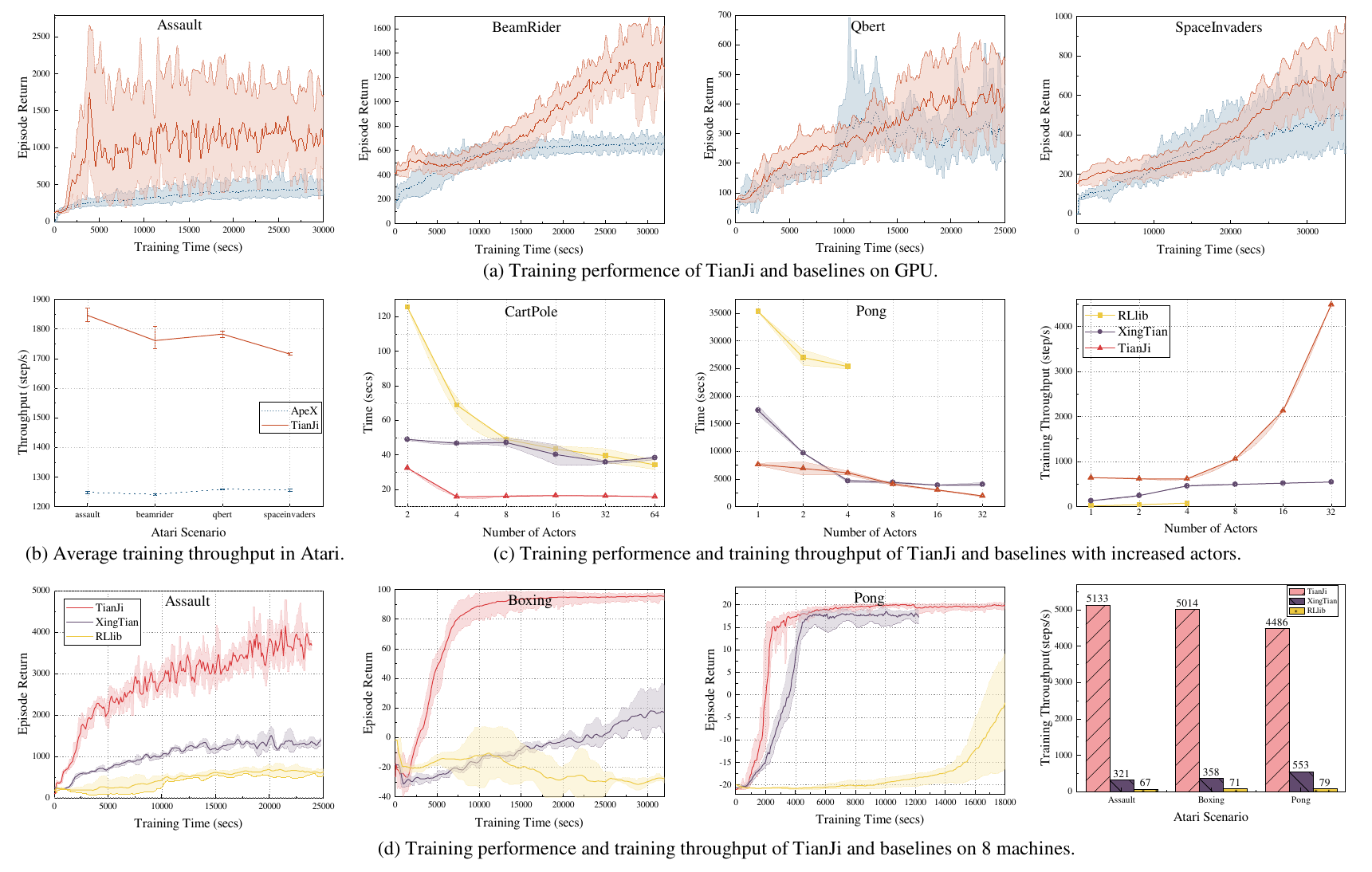} 
	\caption{TianJi outperforms the baselines on learning performance and computational efficiency.}
	\vspace*{-15pt}
	\label{fig4}
\end{figure*}

\subsection{Overall Performance Comparison}
ApeX relaxes certain assignment dependencies and is the most similar work to TianJi. Since ApeX does not support multi-core processing or multiple learners, experiments are conducted on a single GPU machine. Figure \ref{fig4}(a) illustrates the episode return over time, highlighting the learning performance of each system.
The current episode return is derived from clipped rewards within a single life.
TianJi demonstrates superior learning performance compared to ApeX. TianJi’s distributed strategy, based on production-consumption balance, corrects the sample distribution and ensures. In contrast, ApeX still encounters assignment dependencies due
to the data consistency requirements of PER. Conversely, TianJi achieves true relaxation of assignment dependencies, significantly enhancing computational efficiency, as shown in Figure \ref{fig4}(b). ApeX’s open-source implementation is nonscalable, restricting comparisons to a single node. Theoretically, PER’s negative impact on computational efficiency worsens with an increasing number of components. 

Additional performance comparisons were carried out on eight CPU machines, as illustrated in Figure \ref{fig4}(c). For on-policy method learning Cartpole, we compared the final time to reach an average reward of 300. TianJi outperformed the baselines with varying numbers of actors. Compared to RLlib, TianJi achieved a final time speedup of up to 4.36. For off-policy method learning Pong, we compared the final time and training throughput. When the number of actors exceeds 4, XingTian’s final time tends to stabilize. This occurs because the performance bottleneck shifts to training, and increasing sampling efficiency does not improve convergence. The training throughput data show that TianJi’s distributed strategy effectively addresses performance bottlenecks. Compared to XingTian, TianJi achieves 1.6-fold faster convergence and 7.13-fold higher computational efficiency. The experimental results demonstrate that TianJi achieves several times higher training throughput and faster convergence compared to the baselines.
Figure \ref{fig4}(d) provides a detailed performance comparison across eight computational nodes, where episode returns are calculated from game rewards within a single life. TianJi exhibits markedly superior learning performance compared to XingTian and RLlib while achieving significantly higher throughput.

\subsection{Ablation Study}

\subsubsection{Data Transmission Efficiency}
We conducted a virtual communication experiment to evaluate the data transmission efficiency of TianJi and other systems. The experiment,
which followed the communication pattern of a standard DistRL algorithm, concluded after collecting 10,000 samples, with collection time recorded. As a baseline, we selected XingTian, which offers an asynchronous communication channel and exhibits superior data transmission efficiency compared to other systems. We evaluated the data
transmission efficiency by measuring the time to receive the 10,000 samples(referred to as collection time) and the data transmission throughput(referred to as throughput).

\begin{figure}[t]
	\centering
	\includegraphics[width=0.75\columnwidth]{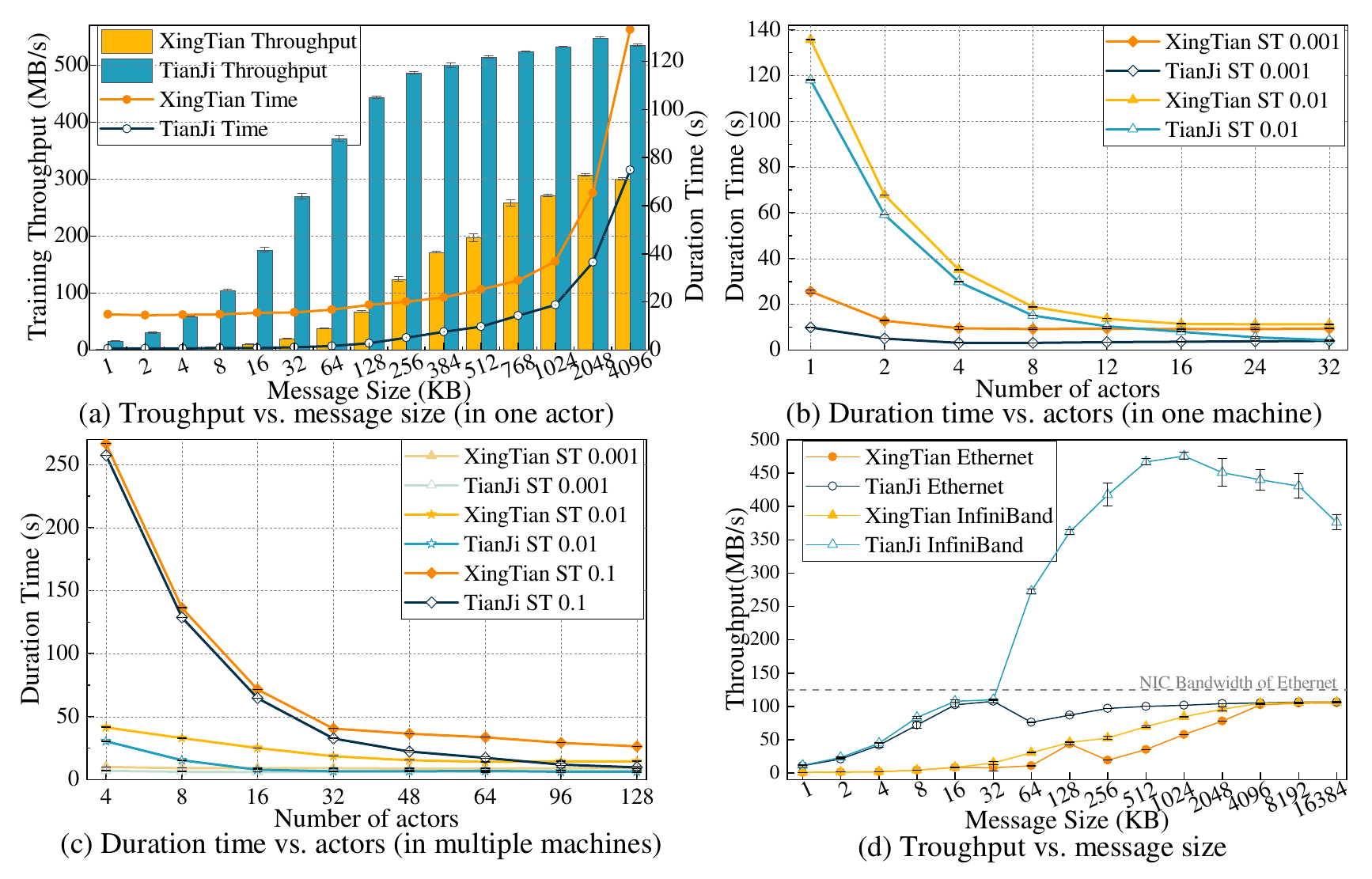} 
	\caption{Comparison of data transfer efficiency between TianJi and XingTian. ST is "sample time".}
	\vspace*{-15pt}
	\label{fig5}
\end{figure}

Figure \ref{fig5}(a) shows that TianJi consistently outperforms XingTian in data transmission efficiency for all message sizes under single-node and single-actor conditions. Figure \ref{fig5}(b) depicts how collection time varies with the number of actors while the message size remains fixed at 512 KB. The figure indicates that, for a fixed sample time(e.g., 0.001),
communication bottlenecks occur with more than four actors, and increasing the number of actors beyond this does not reduce collection time(see Section 3.2.2 for analysis). With a sample time of 0.01, the communication bottleneck occurs with 16 actors. TianJi consistently exhibits lower sample collection times than XingTian across all conditions. Notably, TianJi has a lower communication bottleneck than
XingTian. After reaching the bottleneck, the collection time
depends only on receiver operations(e.g., local storage),
and TianJi’s reception time remains lower.Figure \ref{fig5}(c) shows
how communication efficiency varies as the number of actors increases across multiple computing nodes, with actors evenly distributed among 4 nodes. TianJi consistently has lower collection times than XingTian in all scenarios. Figure \ref{fig5}(d) shows that TianJi achieves throughput close to Ethernet bandwidth. Testing was conducted between two nodes with
Ethernet bandwidth of 125 MB/s and InfiniBand(IB) bandwidth of 12.5 GB/s. Under Ethernet conditions, both TianJi and XingTian show similar performance trends: as message size increases, throughput gradually rises and approaches Ethernet bandwidth. In contrast, under IB conditions, TianJi exceeds previous throughput limits, while XingTian does not. Throughput decreases after 2048 KB due to increased
message generation time with larger message sizes. TianJi
reaches the Ethernet bandwidth limits, demonstrating superior communication efficiency compared to XingTian.

\subsubsection{The Impact of the Distributed Strategy on Convergence}
TianJi proposes a distributed strategy based on production consumption balance to correct sample distribution and maximize training throughput. This section examines the impact of sample staleness and computational-resource mapping on training. Training was conducted on a single node with PPO. The optimal computational-resource mapping, identified by the distributed strategy, includes 2 learners and 4 actors, using 16 cores, denoted as L2A8-C16. A random computational-resource mapping, labeled L1A14-C16, was also evaluated. After introducing asynchrony, simulations with serial sample distributions were conducted using two sample ratios: 1:1 (denoted as New) and 1:8 (denoted as Staleness). Four sets of control experiments were conducted. The results, shown in Figure \ref{fig6}(d), indicate that optimal computational-resource mapping and production consumption balance significantly accelerate computational efficiency and enhance learning performance.

\subsubsection{Scaling Beyond Performance Bottlenecks}
\begin{figure}[tbp]
	\centering
	\includegraphics[width=0.75\columnwidth]{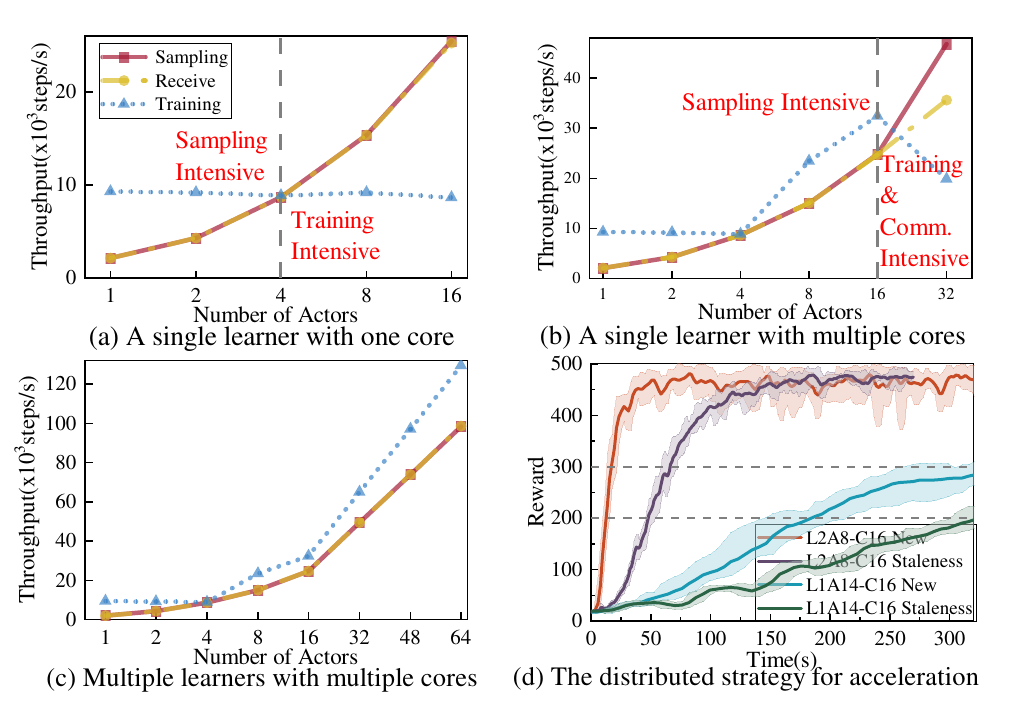} 
	\caption{TianJi’s distributed strategy achieves continuous
		scaling beyond performance bottlenecks and accelerate convergence by adjusting the sample distribution.}
	\vspace*{-15pt}
	\label{fig6}
\end{figure}

Extensive experiments show that once training hits a performance bottleneck, adding resources beyond this point does not improve training performance. Predicting the location of this bottleneck is challenging because various algorithms, applications, and hardware configurations can create bottlenecks in different areas. This section will
demonstrate how TianJi uses effective distribution methods to tackle different performance bottlenecks. TianJi achieves accelerated convergence beyond performance bottlenecks and scales training throughput to near-linear levels. Performance bottlenecks can be categorized into three types: sample-intensive (\ding{192}\ding{193}), training-intensive (\ding{194}), and communication-intensive. In the DQN(Cartpole) experiment, Figure \ref{fig6}(a) illustrates how sampling throughput, receiving throughput, and training throughput change as the
number of actors increases in the DQN(Cartpole) experiment. The number of learners, resources, and computational load is kept constant, resulting in stable training throughput(blue dashed line). As the number of actors increases, sampling throughput(red solid line) also rises. The receiving throughput(yellow dashed line) aligns with the sampling throughput, indicating that no communication bottleneck is
present at this stage. When the number of actors exceeds 4, the learner’s training throughput becomes the bottleneck, and adding more actors does not accelerate performance. Figure \ref{fig6}(b) shows that increasing the number of physical cores for the learners and proportionally enlarging the batchsize can overcome the previous training bottleneck, shifting the equilibrium point to 16 actors. Further increasing the number of actors raises production throughput, but receiving and training throughput fall below this level. This indicates that both communication and training bottlenecks have been
reached. Replicating the equilibrium configuration shown in Figure \ref{fig6}(b) and expanding it in groups, as depicted in Figure \ref{fig6}(c), shows that continuing to increase the number of actors results in a linear increase in sample receiving and training throughput. TianJi employs performance analysis and a distributed strategy based on production-consumption balance to dynamically allocate computation to resources, allowing for sustained expansion beyond performance bottlenecks.

\section{Conclusion}
We found that assignment dependencies are the key to ensuring the equivalence for paralleling DRL training, but also the bottleneck that hinders the performance. Therefore, we introduce TianJi, which reduces assignment dependencies between subtasks using a decentralized, data-driven training approach combined with event-driven asynchronous communication. Additionally, TianJi proposes a distributed strategy based on balancing sample production and consumption to alleviate the convergence issue introduced by relaxing the dependencies. Experimental results show that relaxing assignment dependencies and improving sample quality significantly enhance computational and training efficiency.
TianJi achieves a 1.6-fold speedup in convergence time and a 7.13-fold speedup in training throughput compared to XingTian when scaled to eight machines. The ablation study demonstrates the effectiveness of key components. In future work, we will further deepen the theoretical proof of convergence under the relaxed assignment dependencies.

\section{Acknowledgments}
This work is sponsored in part by the National Natural Science Foundation of China under Grant No.62421002. We extend our gratitude to Peilin Lu and Ruihan Li for their significant contributions to the experiments.

\bibliographystyle{plainnat} 
\bibliography{tianji}     
\newpage 

\appendix
\begin{appendices}
	\section*{Appendix}
	\textbf{Section A} examines the design principles of distributed reinforcement learning(RL) systems. An ideal system should exhibit characteristics such as usability, programming flexibility, versatility, and high performance. \textbf{Section B} discusses the advantages of the TianJi regarding usability and programming flexibility. \textbf{Section C} shows TianJi's versatility , particularly its capability to support various algorithms, applications, and computational platforms. Since high performance is covered extensively in the main paper, \textbf{Section D} analyzes the impact of distributed strategies on learning performance. Finally, \textbf{Section E} presents the hyperparameters used in the experiments.
	
	\section{the Design Principles of Distributed RL System}\label{app:A}
	With the increasing demand for computational power in deep reinforcement learning(DRL), accelerating training has become a research hotspot. Developing effective software platforms is essential for advancing research in this field. Distributed DRL systems can effectively use substantial computational resources to train large-scale data and models. However, what kind of code is useful for research? The deep learning community has identified several key practices, such as modular design, flexibility, usability, performance optimization, visualization, and comprehensive logging. In contrast to deep learning, RL is more irregular. Consequently, there is no consensus on developing DRL system at present.
	Different research objectives result in various trade-offs in software design. Users of DRL system generally fall into two categories: 1) foundational research, which focuses on advancing algorithms; and 2) application deployment, which involves training agents on computational devices. Although researchers and application deployers may prioritize different software requirements, they both demand \textbf{usability}, \textbf{programming flexibility}, \textbf{versatility }, and \textbf{high performance}. Therefore, an effective distributed DRL system should possess these characteristics to address a wide range of use cases.
	
	\section{Usability and Programming Flexibility}
	As agents evolves to handle more complex interactions, creating reusable software for DRL research has become increasingly challenging. TianJi addresses this challenge by balancing usability with programming flexibility and providing standardized, streamlined interfaces. To improve the user experience, TianJi separates user interactions from internal system details. Users can initiate training using configuration files without needing to understand the specifics of the system’s internal implementation. For custom extensions, TianJi offers base class templates for computational and communication tasks. Users can implement custom logic by rewriting the relevant functions in these templates.
	\subsection{Usage}
	In terms of usability, TianJi achieves a high level of abstraction and separation in user experience. This design enables users to start training tasks solely through configuration files, without delving into or understanding the complexities of the underlying implementation. This approach simplifies the workflow and significantly lowers the barrier to entry, allowing even users with limited programming experience to easily conduct model training. By separating user interactions from internal details, the system significantly enhances development efficiency.
	
\begin{figure*}[t]
		\centering
		\includegraphics[width=0.55\textwidth]{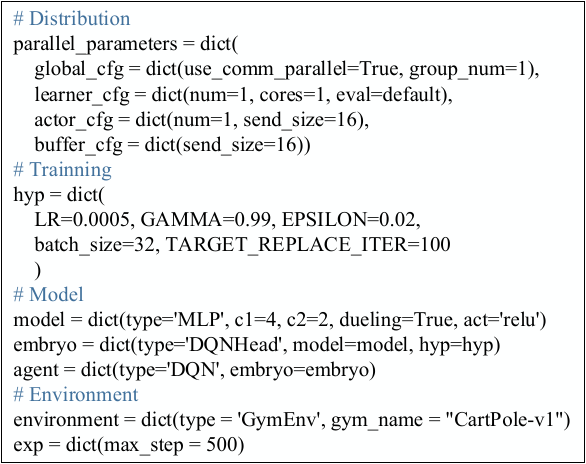} 
		\caption{Configuration File for DQN applied to CartPole task.}
		\label{figa1}
	\end{figure*}

	As shown in Figure \ref{figa1}, the configuration file includes several categories of parameters: distribution parameters, training parameters, model parameters, and environment parameters. Each category provides a comprehensive set of options, enabling users to configure the system flexibly according to specific needs. Distribution parameters allow users to adjust component and inter-component computations, such as the number of computational tasks, computational resources, and number of rollouts. Model parameters offer fine-grained control over model architecture and hyperparameter settings, while environment parameters allow the selection of different runtime environments and their associated settings. By offering an extensive range of parameter settings, the system meets the versatility requirements of TianJi across various application scenarios.
	
	\subsection{Custom Extensions}
\begin{figure*}[t]
		\centering
		\includegraphics[width=0.35\textwidth]{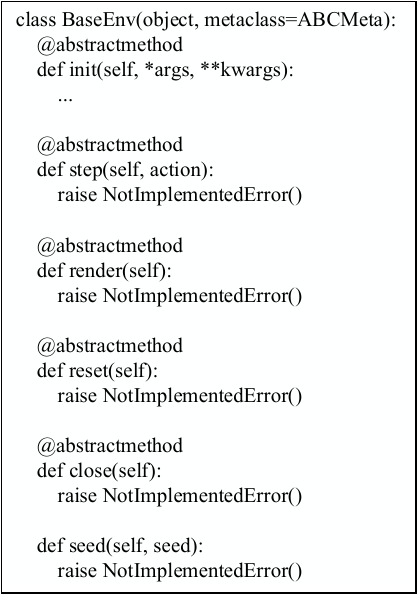} 
		\caption{Template for customizing and extending RL environments.}
		\label{figa2}
	\end{figure*}

	RL practitioners are often not system engineers and may be unfamiliar with mapping computation flows between software and hardware. To address this, TianJi provides a non-intrusive method for implementing custom algorithms, offering significant convenience to users. TianJi features clearly defined components, enabling users to create custom extensions by modifying specific templates. For example, extending a custom task environment is straightforward; users only need to rewrite the relevant functions in the predefined templates (as shown in Figure \ref{figa2}) without altering other parts of the code. This design lowers the barrier to developing custom functionalities, allowing more RL practitioners to focus on algorithm development without being bogged down by underlying system implementation details.
	
	\section{Versatility}
	TianJi offers excellent programming flexibility, supporting a wide range of algorithms, applications, and computational platforms. TianJi currently implements various algorithms, including on-policy, off-policy, and multi-agent reinforcement learning (MARL) algorithms. It supports multiple application scenarios, such as classic control, Atari, Multi-Agent Particle Environment (MPE), and StarCraft II Micro-RTS (SMAC). Additionally, TianJi supports a variety of computational platforms, including pure CPU, pure GPU, and heterogeneous CPU-GPU setups, on single machines as well as across multiple machines.
	
	\subsection{On-policy Algorithm plays Atari games}
\begin{figure*}[t]
		\centering
		\includegraphics[width=0.8\textwidth]{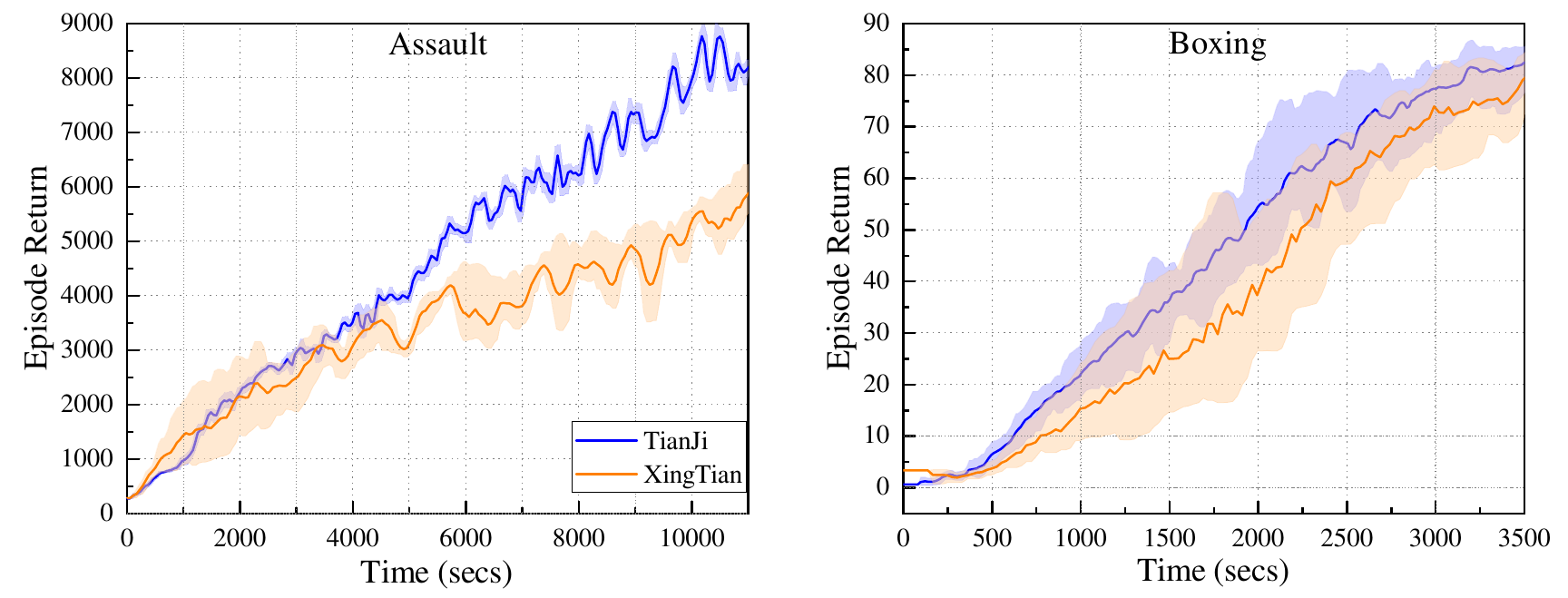} 
		\caption{Learning performance of PPO across different Atari games.}
		\label{figa6}
	\end{figure*}

	TianJi exhibits excellent performance in training agents for both on-policy and off-policy algorithms. The selected algorithm is Proximal Policy Optimization (PPO), which is an on-policy algorithm.  Figure \ref{figa6} illustrates the changes in episode returns over time for PPO. The PPO agents in TianJi were trained using a synchronous mode. Despite this setup, TianJi consistently outperforms the baselines in learning performance. These enhancements are attributed to optimizations in communication and implementation.

	\subsection{Multi-Agent Challenge}
\begin{figure*}[t]
		\centering
		\includegraphics[width=0.77\textwidth]{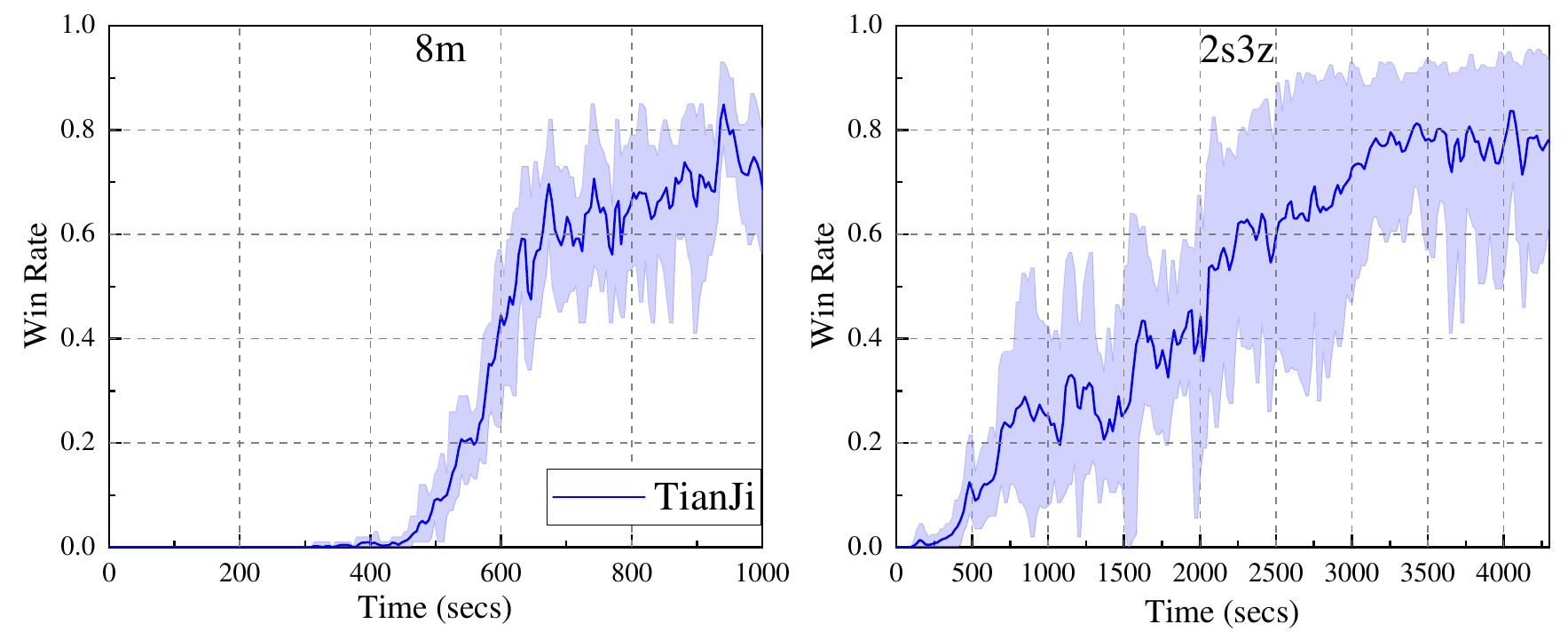} 
		\caption{Learning performance of QMIX across different SMAC maps.}
		\label{figa3}
	\end{figure*}

	TianJi supports both cooperative and competitive multi-agent learning. For example, the SMAC environment is a widely used platform for cooperative MARL, based on Blizzard's StarCraft II real-time strategy game. The objective is for allied agents to collaborate to defeat all enemy units and win the game. We chose QMIX for testing, and Figure \ref{figa3} shows the improvement in agents' win rates throughout the training.

	\begin{figure*}[t]
		\centering
		\includegraphics[width=0.8\textwidth]{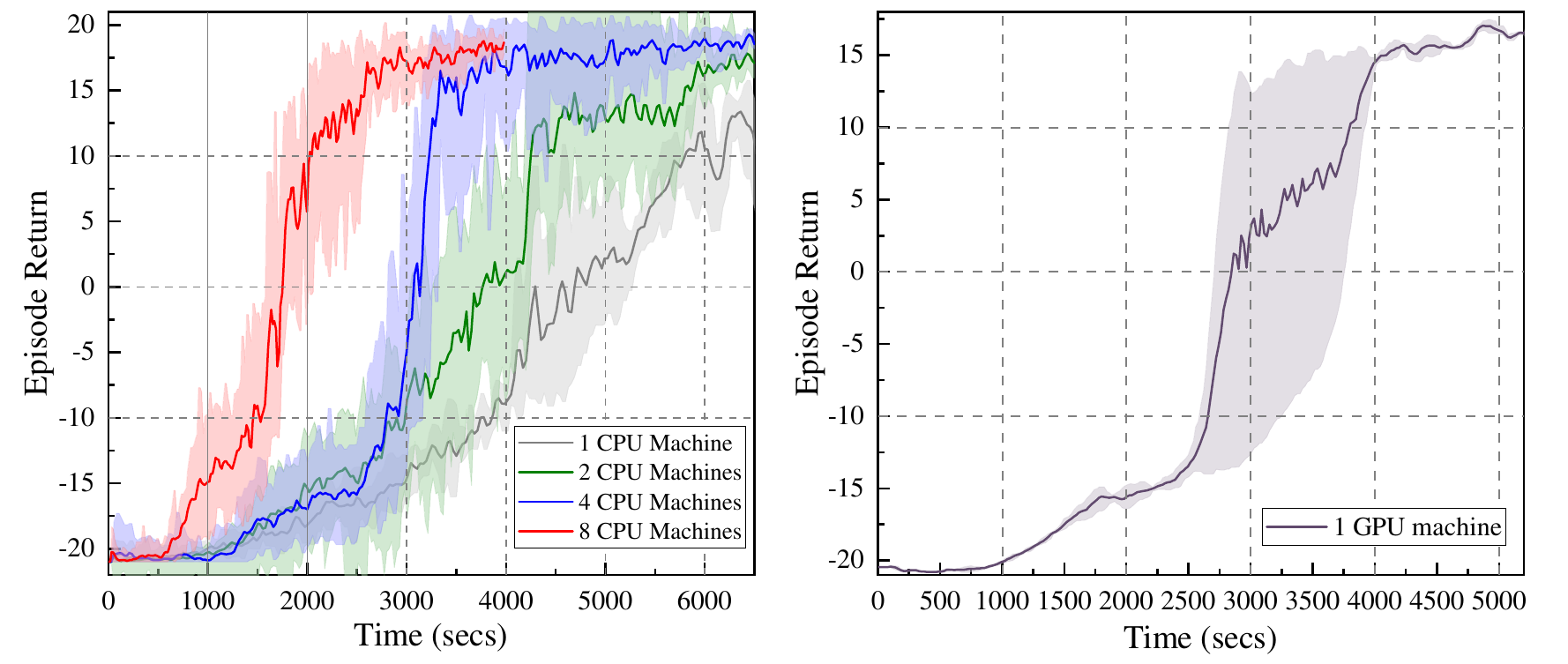} 
		\caption{Performance of TianJi training on 8 CPU machines and single GPU machine.}
		\label{figa4}
	\end{figure*}
	
	\subsection{Support for Multiple Computing Platforms}
	TianJi supports various computational platforms, including pure CPU, pure GPU, and heterogeneous CPU-GPU setups.The system efficiently performs training tasks in both single-machine and multi-machine setups. In this section, we compare the training process of the same algorithm and application across different computational platforms, including training on eight machines with pure CPU and single-machine training with GPU. Figure \ref{figa4} shows the experimental results for these two configurations.

	\section{Further Discussion on Distributed Strategies}
	TianJi introduces a distributed strategy based on balancing sample production and consumption. This strategy addresses sample quality issues and overcomes performance bottlenecks in scaling. Section D.1 examines how sample staleness affects learning efficiency, demonstrating that balancing sample production and consumption effectively corrects sample distribution. Section D.2 discusses whether accelerating processes outside performance bottlenecks improves learning efficiency. This distributed strategy, grounded in performance analysis and the principle of sample production-consumption balance, corrects sample quality issues and overcomes performance bottlenecks, ensuring convergence in training.
	\subsection{Sample Staleness}
	This section discusses the impact of sample staleness on learning efficiency. Proximal Policy Optimization (PPO) is an on-policy algorithm in which sample production and consumption throughputs are equal. Thus, sample staleness can be controlled by adjusting the buffer size. When the batch size equals the buffer size, all samples are new, referred to as New. When the buffer size is twice the batch size, the samples consist of half new and half old samples, referred to as \textit{Lag2}, and so on. As shown in Figure \ref{figa5}.a, sample staleness significantly affects learning efficiency. A higher proportion of old samples results in slower convergence. Thus, in fully asynchronous training, controlling the ratio of old to new samples through a producer-consumer balance to approximate the serial ratio is an effective way to accelerate convergence.
		
	\subsection{Performance Bottlenecks}
\begin{figure*}[h]
		\centering
		\includegraphics[width=0.9\textwidth]{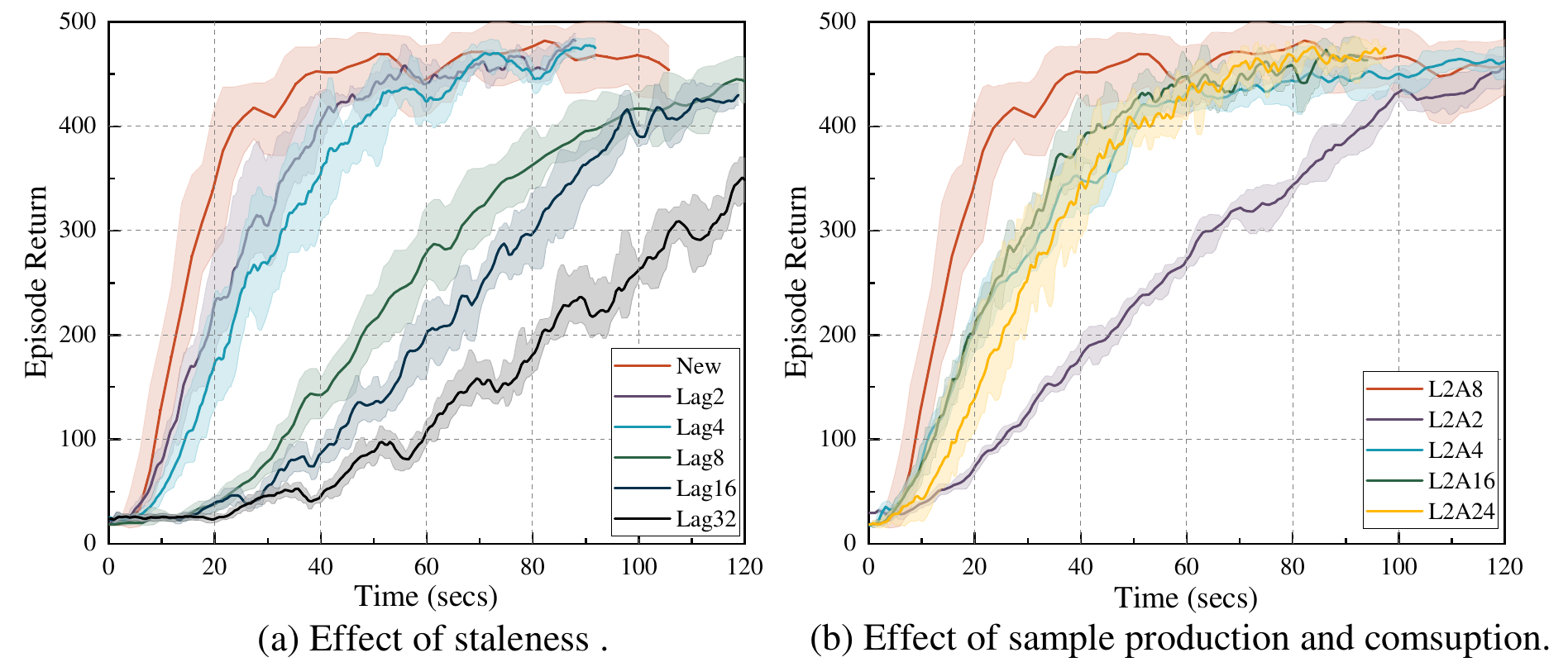} 
		\caption{Impact of sample staleness and computational throughput on training performance.}
		\label{figa5}
	\end{figure*}

	Section D.2 discusses whether accelerating processes outside performance bottlenecks improves learning efficiency. With two fixed learners, each using four cores, the theoretical sample consumption throughput remains constant. As the number of actors increases from 2 to 24, sample production throughput also increases. According to Table 1, at \textit{L2A8}, sample consumption throughput reaches its maximum. \textit{L2A8} indicates a configuration with 2 learners and 8 actors; other configurations follow this pattern. This indicates that at \textit{L2A2} and \textit{L2A4}, the performance bottleneck is due to sample production throughput during exploration. Beyond \textit{L2A8}, the performance bottleneck shifts to sample consumption throughput during training. At this point, increasing the number of actors further enhances sample production throughput but does not improve learning efficiency. In fact, due to factors such as increased interactions, sample consumption throughput may decrease, leading to reduced learning efficiency. Figure \ref{figa5}.b shows learning efficiency for different numbers of actors.Comparing \textit{L2A8} to \textit{L2A4}, mitigating performance bottlenecks in sample production led to improved learning efficiency. However, comparing \textit{L2A16} to \textit{L2A8}, accelerating computations outside the performance bottleneck did not improve learning efficiency. Therefore, addressing performance bottlenecks is crucial for accelerating learning. TianJi’s distributed strategy is designed to achieve this, as detailed in Sections 3.3 and 4.3.3 of the main paper.
	
	\begin{table*}[h]
		\centering
		\resizebox{0.8\textwidth}{!}{
			\begin{tabular}{c|c|c|c}
				\hline
				Setting & Final Time(s) & Production Throughput(steps/s) & Consumption Throughput(steps/s) \\
				\hline
				\multirow{3}{*}{L2A2} & 61.29 & 3255.02 & 3263.15 \\
				& 65.48 & 3293.26 & 3299.35 \\
				& 61.75 & 3280.98 & 3288.98 \\
				\hline
				\multirow{3}{*}{L2A4} & 31.52 & 6232.26 & 6199.11 \\
				& 30.80 & 6224.91 & 6174.51 \\
				& 34.77 & 6229.98 & 6174.51 \\
				\hline
				\multirow{3}{*}{L2A8} & 19.01 & 11661.47 & 11432.79 \\
				& 19.40 & 11571.96 & 11308.52 \\
				& 20.10 & 11578.31 & 11247.72 \\
				\hline
				\multirow{3}{*}{L2A16} & 28.77 & 22137.23 & 10725.61 \\
				& 25.84 & 22335.65 & 10894.60 \\
				& 29.31 & 22176.82 & 10781.47 \\
				\hline
				\multirow{3}{*}{L2A24} & 35.54 & 31190.68 & 10189.90 \\
				& 27.70 & 32251.47 & 10508.93 \\
				& 33.75 & 32814.92 & 10671.10 \\ 
				\hline
			\end{tabular}
		}
		\caption{Changes in training convergence time, sample production, and consumption throughput with increasing number of actors.}
		\label{table1}
	\end{table*}
	
	\section{Hyperparameters}
	Key hyperparameter values from the experiments are shown in Table 2. Among these, the learning rate and batch size vary and are significantly affected by scaling. Due to computational constraints, we could not perform a comprehensive sweep and adjustment of all hyperparameters in separate experiments. Consequently, the parameter values in Table 1 do not represent the optimal configuration but are provided for reference. To ensure fairness in comparative experiments, we kept consistent parameter settings across different algorithms for the same task. This consistency helps eliminate the influence of other variables, ensuring that the experimental results accurately reflect performance differences between algorithms.
	
	\begin{table*}[h]
		\centering
		\resizebox{0.85\textwidth}{!}{
			\begin{tabular}{l|l|l|l|l}
				\hline
				Hyperparameter & PPO & DQN(CartPole) & DQN(Atari) & QMIX  \\
				\hline
				discount $\gamma$ & 0.99 & 0.99 & 0.99 & 0.99  \\
				greedy $\lambda$ & - & 0.98 & 0.98  & 0.99 \\
				decay step & - & - & - & 1e4 \\
				gae & 0.95 & - & - & - \\
				target update interval & 10 & 100 & 64 & 64  \\
				hidden layer & 256 & 256 & 512 & 32  \\
				max gradient norm  & - & - & - & 10 \\
				optimizer & adam & adam & adam & prmsprop \\
				optimizer config & $\epsilon=1e-5$ & $\epsilon=1e-8$ & $\epsilon=1e-8$ & $\alpha=0.99, \epsilon=1e-5$ \\	
				learning rate & 1e-4 & 5e-4 & 2.5e-5 & 1e-4 \\
				rollout length & 512 & 16 & 16 & 16 \\
				buffer warmup size & 1 & 32 & 625 & 32 \\
				buffer size & =batch size & 2048 & 25000 & 5000 \\
				\hline
			\end{tabular}
		}
		\caption{hyperparameters used in benchmark experiments (in CPUs).}
		\label{table2}
	\end{table*}
	
\end{appendices}

\end{document}